\documentclass{article} 
\usepackage{makeidx}  
\usepackage{epsfig} 
\usepackage{url}
\usepackage[latin1]{inputenc}
\usepackage{color}
\usepackage{alltt}

\begin{document} 

%
\title{Evolving XSLT stylesheets} 

\author{N\'estor Zorzano, Daniel Merino, JLJ Laredo,\\ 
JP Sevilla, Pablo Garc\'ia, JJ Merelo}
 
\maketitle              
\begin{abstract} 
This paper introduces a procedure based on genetic programming to evolve
XSLT programs (usually called stylesheets or logicsheets). XSLT is a
general purpose, document-oriented functional language, generally used
to transform XML documents (or, in general, solve any problem that can
be coded as an XML document). The
proposed solution uses a tree representation for the stylesheets as
well as diverse specific operators in order to obtain, in the studied
cases and a reasonable time, a XSLT stylesheet that performs the
transformation. Several types of representation have been compared,
 resulting in different performance and degree of success.
\end{abstract} 

\section{Introduction}

Since the IT industry has settled in different XML (eXtensible Markup
Language) \cite{xml:bible} dialects as information
exchange format, there is a business need for programs that transform
from one format to another, extracting information or combining it in
many possible ways; a typical example of this transformation could be the extraction of
news headlines from a newspaper in Internet that uses XHTML\footnote{An XML version of
the Hypertext Markup Language (HTML) used in web pages.}. 

XSLT stylesheets (XML Stylesheet
Language for Transformations) \cite{XSLT}, also called {\em
logicsheets} are designed for this purpose: applied to an XML document, they produce another. There are
other possible 
solutions: programs written in any language that work with text as
input and output,  programs using regular expressions and SAX filters
\cite{wiki:sax}, that process each tag in a XML document in a
different way, and do not need to load into memory the whole XML
document. However, they need external
languages to work, while XSLT is a part of the XML set of standards, and, in
fact, XSLT logicsheets are XML documents, which can be integrated
within an XML framework; that is why XSLT is, if not the most common,
at least a quite usual way of transforming XML documents. 

The amount of work needed for logicsheet creation is a problem that
scales quadratically with the quantity of initial and final formats. For
$n$ input and $m$ output formats, $n \times m$ transformations will be
needed\footnote{If an intermediate language is used, just $n + m$,
but this increases the complexity of the transformation and decreases
its speed.}. Considering that each conversion is a hand-written program
and the initial and final formats can vary with certain frequency, any
automation of the process means a considerable saving of effort on
the part of the programmers. 

The objective of this work is to find the XSLT logicsheet
that, from one or several input XML documents, is able to obtain an output XML
document that contain exclusively the information that is considered
important from original XML documents. This information may be ordered
in any  possible way, possibly in an order different to the input
document. This logicsheet will be evolved using Genetic Programming;
XSLT programs are obviously not written in LISP (or Lisp-like s-expressions), as is
usual in GP, but they also have a tree structure (as any XML
document), since they are represented internally as a DOM (Document
Object Model) tree.  

In order to evolve XSLT logicsheets, we will have to take into account
this structure. XML is an extensible markup language, in other words,
a language that 
allows to define elements (tags) and the grammar that they follow. XML is
based on the concept of encapsulation: document fragments are
encapsulated in an area delimited by two tags. All XML documents have
a tree structure (the so-called Document Object Model --DOM-- tree) with a single root element that
contains(encapsulates) all the contents of the document. In addition,
XML elements  have attributes which contain other information 
needed for the processing of the document. Sometimes, the elements and
attributes have a syntax or semantics determined by Data Type
Dictionary (DTD) or XSchema (equivalent concept that uses XML for its
definition), in which case the document can be validated; however, in
most applications what is called {\em well-formed} XML is more than
enough. 

Thus, XSLT provides a general mechanism for the
association of patterns in the source XML document to
the application of format rules to these elements, but in order to simplify
the search space for the evolutionary algorithm, only three
instructions of XSLT will be used in this work: {\sf template}, which
sets which XML fragment will be included when the element matching its
\textsf{match} attribute is found; {\sf
apply-templates}, which
is used to select the elements to which the transformation is going to
be applied and delegate control to the corresponding {\sf templates};  and {\sf value-of}\footnote{With {\sf text} used for easy
visualization of the final document}, which simply includes the
content of an XML document into the output file. This implies also a simplification of
the general XML-to-XML transformation problem: we will just extract
information from the original document, without adding new elements
(tags) that did not exist in the original document. In fact, this
makes the problem more similar to the creation of an {\em scraper}, or
program that extracts information from legacy websites or
documents. Thus, we intend this paper just as a proof of concept,
whose generalization, if not straightforward, is at least possible. 
 
We also take into account XPath 
\cite{XPath} as a key element within the XML family of standards. XPath defines a way to locate a specific element within a XML
document, by using references to specific nodes in the document, in an
way similar to file access in a file-system tree; in the XPath
specification, a document is considered as a tree,
accessible by position. In addition, XPath provides a way to select groups of
elements ({\em node-sets}) and to filter them by using predicates allowing,
for instance, to select the element that occupies a certain position
within a node-set.

The rest of the paper is structured as follows: the state of the art
is presented in section \ref{sec:soa}. Section \ref{sec:methodology} describes the
solution presented in this work. Experiments are described in  section
\ref{sec:exp}, with the automatic generation of XSLT stylesheets for two examples 
and finally the conclusions and possible lines of future work are
presented in section \ref{sec:conc}.

\section{State of the art}
\label{sec:soa}

So far, very few papers about applying genetic programming techniques to
the 
automatic generation of XSLT logicsheets have been published; one of these, by
Scott Martens \cite{martens00},  presents a technique to find XSLT stylesheets
that transform a XML file into HTML  by using genetic
programming. Martens works on simple XML documents, like the ones shown
in its article, and uses the UNIX diff function as the basis for its
fitness function. He concludes that genetic programming is useful to
obtain solutions to simple examples of the problem, but it needs
unreasonable execution times for complex examples and might not be a
suitable method to solve this kind of problems. However, computing has
changed a lot in the latest seven years, and the time for doing it is
probably now, as we attempt to prove in this paper.

Unaware of this effort, and coming from a completely different field,
Schmidt and Waltermann \cite{schmid04} approached the problem taking
into account that XSLT is a functional language, and using functional
language program generation techniques on it, in what they call {\em
inductive synthesis}. First they create a non-recursive program, and
then, by identifying recurrent parts, convert it into a recursive
program; this is a generalization of the technique used to generate
programs in other programming languages such as LISP
\cite{blisp,summers}, and used thoroughly since the eighties
\cite{Biermann:83}. 

A few other authors have approached the general problem of generating XML
document transformations knowing the original and target structure of
the documents, as represented by its DTD: Leinonen et
al. \cite{Leinonen,Kuikka02} have proposed semi-automatic generation
of transformations for XML documents; user input is needed to define the label
association. There are also freeware programs that perform transformations on
documents from a XSchema to another one. However, they must know both
XSchemata in advance, and are not able to accomplish general
transformations on well formed XML documents from examples. 

The automatic generation of XSLT logicsheets is also a super-set of 
the problem of generating 
{\em wrappers}, that is, programs that extract information from
websites, such as the one described by Ben Miled et al. in
\cite{benmiled:wrappers}. In fact, HTML is similar in structure to XML
(and can actually be XML in the shape of XHTML), but these programs do
not generate new data (new tags), but only extract information already
existing in web sites. This is what applications such as X-Fetch Wrapper,
developed by Republica\footnote{This company no longer exists, and the
product seems to have been discontinued}, do. The company that
marketed it claims that it is able to perform transformation between
any two XML formats from 
examples. Anyway, it is not so clear that transformations are that
straightforward:  according to a
white paper found at their website, it uses a document transformation language.

\section{Methodology}
\label{sec:methodology}

XSLT
stylesheets have been inserted into tree structures, making them
evolve by using variation operators. Each XSLT stylesheet is evaluated
using a fitness function that  shows the adjustment rate between
generated XML and output XML associated to the example. The solution
has been programmed using JEO \cite{jeo-gecco2002}, an evolutionary
algorithm library developed at University of Granada as part of the
DREAM project \cite{LNCS2439:ID197:pp665}, which is available at \url{http://www.dr-ea-m.org} together with the rest of the project.

The generated XML documents are encapsulated within an XML tag whose
name equals the root element from the input XML. Next, structures used for evolution and operators applied to them are
described. These operators work on data structures and XPath queries
within them.

The search space over possible stylesheets is exceedingly large. In
addition, language grammar must be considered in order to avoid
syntactically wrong stylesheet generation. Due to this,
transformations are applied to predetermined stylesheet structures
which have been selected. These transformations alter the structure
and preserve the syntax. This limits search space, generating
suboptimal solutions, so three stylesheets structures that are not
changed by transformations have been selected.  

Next we will describe the three different XSLT stylesheet structures
that have been used in the experiments, and the operators that are
applied on them. 

\subsection { First structure}

\begin{itemize}
\item The XSLT logicsheet will have three levels of depth. First level
is the root element $<$xsl:stylesheet$>$ which is common to all XSLT
stylesheets.  
\item An undetermined quantity of $<$xsl:template match=...$>$ instructions hangs from the root element.
\item The value of match attribute for the first template that hangs
off the root will be ``/''. This template and its content never will
be modified by the apply of operators. The only instruction inside
this element will be apply-templates, that will have a select
attribute whose value will be a ``/'' slash followed by the root
element name. Thus the rest of templates included in the stylesheet
will be processed. 
\item The values for the match attributes for the rest of the templates
      from the second will be simply tag names of the input XML. Every
      value will have an undetermined number of children, that will be
      apply-templates or value-of instructions. These instructions will
      have select attributes, whose values will be XPath relative
      routes, built over the template path. Those routes would include
      every possible XPath clauses. {\sf value-of} will be used instead
      of {\sf apply-templates} if the when the value is self ({\tt .}).
\end{itemize}

\subsection{Second and thirds structure (types 2 and 3)}
The main differences with the first one are:
\begin{itemize}

\item The value of the match attribute for the first template that
hangs off the root will be ``/'' too, but, in this case it will have
an indeterminate number of children, that will be all apply-templates
instructions, whose values  for the select attribute will be XPath
absolute valid routes in the input XML, that will include only tag
names separated to each other by a unique slash. 

\item The values for the match attributes for the other templates that
      hang from the XML root
will be the same values that had the select attributes of the
{\sf apply-templates}  in the first template. Therefore, there will be
      as many {\sf template} instructions as
      the number of {\sf apply-templates} in it, and they will be located in the same order.

\item Every template of the previous section will have an undetermined
number of children, and all of them will be value-of instructions,
where the value for the select attribute will be XPath routes relative
to the XPath absolute route of the father template. These routes would
include every mechanisms of XPath that the designed operators  allow. 

\item If the absolute route of a template has a maximum depth level
inside the XML structure, its only \textsf{value-of} child will have
      {\sf select} the self element: ``.''. 

\end{itemize}

Type 3 structure  is identical to the previous one, but the children of the
template instructions will be \textsf{apply-template} instead of
\textsf{value-of} instructions, except when the XPath of the select attribute
is ``.''. This structure could be used when the two previous
structures do not yield good results.

\subsection{Genetic operators}

The operators may be classified in two different types: the first one
consists in operators that are commons to the three structures and whose
assignment is to modify the XPath routes that contains the attributes of
the XSLT instructions (specially apply-template and value-of). Operators in the second group are used to modify the XSLT
tree structure and take different shape in each of them (so that the structure is kept). In order to ensure the
existence of the elements (tags) added to the XPath expressions and XSLT instruction attributes, every time one of them is needed it is randomly
selected from the input file. 

The common operators are:

\begin{itemize}
\item {\sf XSLTreeMutatorXPath(Add$|$Mutate$|$Remove)Filter}: Adds, changes
      number, or removes a cardinal
      filter to any of the XPath tags that allow it. For example:
      \texttt{/book/chapter $\rightarrow $ /book/chapter[4],
      /book/chapter[2] $\rightarrow $ /book/chapter[4], /book/chapter[2]
      $\rightarrow $ /book/chapter}. 

\item {\sf XSLTTreeMutatorXPathAddBranch}: Adds to a XPath route a new
      tag, chosen randomly from the possibles, observing the hierarchy
      of the input XML file tree: {\tt /book/chapter $\rightarrow $
      /book/chapter/title }

\item {\sf XSLTTreMutatorXPathSetSelf}: Replaces the deepest node tag of a XPath route by the self node.

\item {\sf XSLTTreeMutatorXPathSetDescendant}: Removes one of the
      intermediate tags from a XPath route, remaining a Descendant type
      node: \texttt{/book/chapter/title $\rightarrow$ /book//title}.

\item {\sf XSLTTreeMutatorXPathRemoveBranch}: Removes the deepest
  element tag of a XPath route, ascending a level in the XML tree. For
  example: {\tt /book/chapter/title $\rightarrow $ /book/chapter}.

\end{itemize}

The operators that change the DOM structure of the XSLT logicsheet are:

\begin{itemize}

\item {\sf XSLTTreeCrossoverTemplate}: Swaps template instructions
      subtrees between the two {\em parents}. This is the only
      crossover-like operator.

\item {\sf XSLTTreeMutator(Add$|$Mutate$|$Remove)Template}: Inserts, changes
      or removes a template. Insertion is performed on the root element matching an random element. The choice of
      this random element gives more priority to the less deeper
      tags. The position of the new template inside the tree will be
      randomly selected, and its content will be
      \texttt{apply-templates} or \texttt{value-of} tags with the select
      attribute containing XPath routes relatives to the parent template
      XPath route randomly generated using the XPath operators. Change
      operates on a random node, generating a new subtree; and removal
      also eliminates a random template (if there are more than two).
 
 \item {\sf XSLTTree(Add$|$Remove)Apply}: It adds or removes a child to a randomly
selected template present in the tree. The position of the new leaf
inside the subtree that represents the template also will be randomly
selected. The new element is randomly generated from the route that
contains its parent template instruction. The \textsf{Remove} operator
      also deletes the template node if the removed child was the last
      remaining one, but it is not applied if there is a single template left.

 \item {\sf XSLTreeMutateApply(1$|$2)}: Changes a randomly selected child
      (1) or creates a relative XPath from the one that contains the
      father XSLT:template and the XPath of the leaf that we are going
      to modify (2).
 
\item {\sf XSLTreeSetTemplateNull}: It chooses a subtree template from
the XSLT tree and replaces its content by a single instruction
$<$xslt:value-of select=''.''$>$.  In the cases of second and third XSLT
tree structure, there are nine equivalent operators.

\end{itemize}

\subsection{Fitness function}

Each XSLT sheet is applied to a XML input file and evaluated by comparing the 
result with the objective XML document. The evaluation function for
each individual solution is represented next:

\begin{equation}
F = \frac{D}{L_1}+ \Big(\frac{S}{2}\Big)^2 + \frac{L_2}{10000} 
\end{equation}

Where:
\begin{itemize}
\item {\bf $D$} represents the number of lines where the processed and
objective XML documents differ.
\item {\bf $L_1$} is the number of lines in the obtained XML.
\item {\bf $L_2$} is the number of lines of the XSLT.
\item {\bf $S$} corresponds to:
\begin{itemize}
\item 0 when the number of lines in the objective XML - $L_1$ $<$ 0.
\item $L_1$ - number of lines in the objective XML otherwise.
\end{itemize}
\end{itemize}

By taking into account not only the difference between the desired and
obtained XML but also parameters related to the size of the resulting
XSLT stylesheet; that way, there is a selective pressure for more
compact programs, in an attempt to avoid bloating and useless
structures. 


\section{Experiments and results}
\label{sec:exp}

To test the algorithm we have performed several experiments with different
XML input files and an unique XML output file. The algorithm has been
executed five times for each input XML and for each of the three XSLT
structures shown in the previous section. 

\begin{figure}
\input {highlight.sty}
\noindent
\ttfamily
\small
\hlstd{}\hlkwa{$<$?xml version=}\hlstr{"1.0"}\hlkwa{ encoding=}\hlstr{"ISO{-}8859{-}1"}\hlkwa{?$>$}\hlstd{}\hspace*{\fill}\\
\hspace*{\fill}\\
\hlkwa{$<$biblioteca\textunderscore musical$>$}\hlstd{\hspace*{\fill}\\
	}\hlkwa{$<$disco$>$}\hlstd{\hspace*{\fill}\\
}\hlstd{\ \ }\hlstd{}\hlkwa{$<$titulo$>$}\hlstd{I}\hlkwa{$<$/titulo$>$}\hlstd{\hspace*{\fill}\\
}\hlstd{\ \ }\hlstd{}\hlkwa{$<$autor$>$}\hlstd{Led Zeppelin}\hlkwa{$<$/autor$>$}\hlstd{\hspace*{\fill}\\
}\hlstd{\ \ }\hlstd{}\hlkwa{$<$cancion$>$}\hlstd{God Times, Bad Times}\hlkwa{$<$/cancion$>$}\hlstd{\hspace*{\fill}\\
}...\\
\hlstd{\ \ }\hlstd{}\hlkwa{$<$fecha\textunderscore grabacion$>$}\hlstd{\hspace*{\fill}\\
}\hlstd{\ \ \ }\hlstd{}\hlkwa{$<$mes$>$}\hlstd{Mayo}\hlkwa{$<$/mes$>$}\hlstd{\hspace*{\fill}\\
}\hlstd{\ \ \ }\hlstd{}\hlkwa{$<$mes$>$}\hlstd{Junio}\hlkwa{$<$/mes$>$}\hlstd{\hspace*{\fill}\\
}\hlstd{\ \ \ }\hlstd{}\hlkwa{$<$ano$>$}\hlstd{}\hlnum{1969}\hlstd{}\hlkwa{$<$/ano$>$}\hlstd{\hspace*{\fill}\\
}...\\
\hlkwa{$<$/biblioteca\textunderscore musical$>$}\hlstd{}\hspace*{\fill}\\
\mbox{}
 \caption{Part of the first XML document used for experiments.\label{fig:xml1}}
\end {figure}

The first input XML file (see figure \ref{fig:xml1}) is a document that describes musical records from diverse authors, from which we want to extract the name of the
authors and the first song of their disk, while the second XML input file includes one extra disk
whose songs and author are not present in the XML output file.

The computer used to perform the experiments is a Centrino Core Duo at
1.83 GHz, 2 GB RAM, and the Java Runtime Environment 1.6.0.01;
the termination condition was set to 100 generations or until a solution was found and the selector
was a Tournament selector with 5 individuals;
5 experiments were run, with different random seeds, for each template
type and input document. The parser libraries used have been Xalan 2.7.0 for XSLT and
Xerces 2.8.1 for XML.
 The parameters used in the experiments, and which were set to a default
 value with no attempt to optimize them, are
 shown in table \ref{tab:priorities}; results are shown in tables
 \ref{tab:doc1} and  \ref{tab:doc2}.

\begin{table}[htb]
\begin{center}
  \begin{tabular}{lr}
   \hline
   Operator & Priority \\
	\hline
XSLTTreeMutatorXPathSetSelf & 0.10\\
XSLTTreeMutatorXPathSetDescendant & 0.24 (Only Type 1)\\
XSLTTreeMutatorXPathRemoveBranch & 0.27 (Type 2-3) 0.39 (Type 1)\\
XSLTTreeMutatorXPathAddBranch & 0.99 \\
XSLTTreeMutatorXPathAddFilter & 0.45 (Type 2-3) 0.53 (Type 1)\\
XSLTTreeMutatorXPathMutateFilter & 0.64 (Type 2-3) 0.69 (Type 1)\\
XSLTTreeMutatorXPathRemoveFilter & 0.83\\
   \hline 
   XSLTTreeCrossoverTemplate & 0.61 (Type 1), 0.11 (Types 2 and 3) \\ 
   XSLTTreeMutatorAddTemplate & 0.13 \\
   XSLTTreeMutatorMutateTemplate & 0.11\\ 
   XSLTTreeMutatorRemoveTemplate   & 0.13\\ 
   XSLTTreeAddApply       & 0.11\\ 
   XSLTTreeMutateApply1      & 0.11\\ 
   XSLTTreeMutateApply2      & 0.11\\ 
   XSLTTreeRemoveApply       & 0.15\\ 
   XSLTTreeSetTemplateNull   & 0.04\\ 
  \end{tabular}
 \caption{Operator priorities (used for the roulette wheel that randomly
 selects the operator to apply) used in the
 experiments.\label{tab:priorities} }
\end{center}
\end{table}
\begin{figure}[htb]
  \centerline{
  \begin{tabular}{cc}
  \includegraphics[width=6cm,height=5cm,clip=]{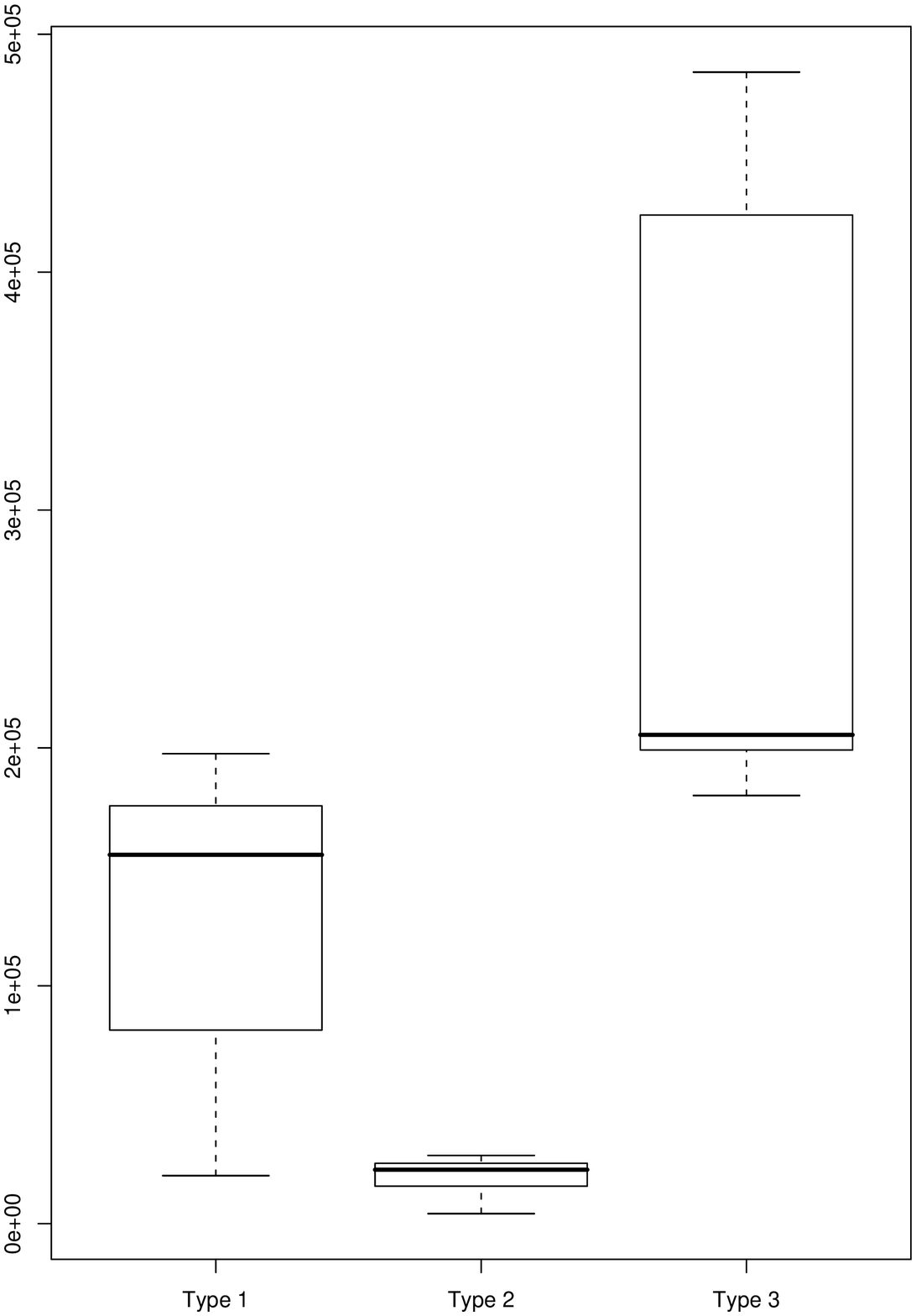} & 
  \includegraphics[width=6cm,height=5cm,clip=]{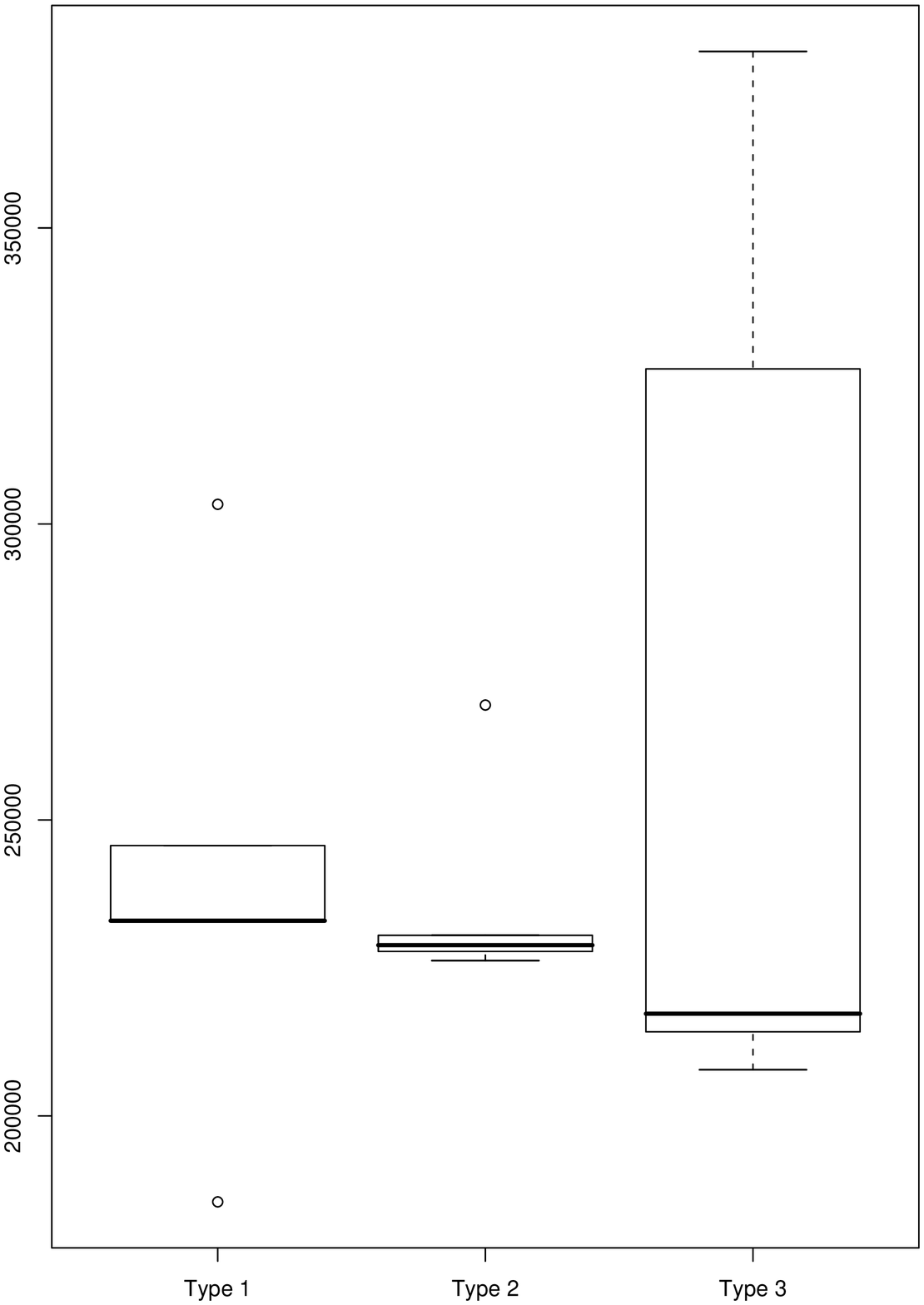}\\
  \end{tabular}
  }
\caption{Distribution of running time (100 generations, or until a
 solution is found) for the first input document
 (left) and the second (right), in milliseconds. In the first case, Type
 2 algorithm reaches the solution more frequently than the others, thus
 beating them in running time. However, in the second case, solution is
 not found during the 100 generations it was allowed to run; even so,
 Type 2 is consistently faster than the others. }
\label{fig:time}
\end{figure}
\begin{table}[htbp]
\small
\begin{center}
\begin{tabular}{|l|c|c|c|}
\hline
 & Success rate & Time & Number of generations \\
\hline
 Type 1  &  0.6   &   125968 $\pm$73509  &  66 $\pm$ 40 \\
 Type 2  &  0.8   &   19359 $\pm$ 9709  &  6 $\pm$ 4  \\
 Type 3  & 0      &  298528 $\pm$ 143834   &  100 \\
\hline
\end{tabular}
\end{center}
\caption[]{Results for the first input document; success rate is the
 number of times a solution is reached within the allotted 100
 generations; running time is in milliseconds, and the number of
 generations needed to find the solution within the 100 allotted generations.
}
\label{tab:doc1}
\end{table}
These experiments assess the ability of the different evolution models
 to find a solution in a simple (document 1) and a slightly more complicated case (document 2). In the
 first case, it is not too difficult to find the solution in a few
 generations, but Type 2 templates are more successful than the rest,
 finding the solution in most cases, and doing so in less generations
 (thus, less evaluations); the XSLT logicsheet found is shown in figure \ref{fig:sol}. Type 3 is never able to find a solution in
 the given time. 
\begin{figure}[htb]
\input {highlight.sty}
\noindent
\ttfamily
\small
\hlstd{}\hlkwa{$<$?xml version=}\hlstr{"1.0"}\hlkwa{?$>$}\hlstd{}\hspace*{\fill}\\
\hlkwa{$<$xsl:stylesheet version=}\hlstr{"1.0"}\hlkwa{ xmlns:xsl=}\hlstr{"http://www.w3.org/1999/XSL/Transform"}\hlkwa{$>$}\hlstd{\hspace*{\fill}\\
}\hlstd{\ \ }\hlstd{}\hlkwa{$<$xsl:output indent=}\hlstr{"no"}\hlkwa{ method=}\hlstr{"xml"}\hlkwa{/$>$}\hlstd{\hspace*{\fill}\\
}\hlstd{\ \ }\hlstd{}\hlkwa{$<$xsl:template match=}\hlstr{"/"}\hlkwa{$>$}\hlstd{\hspace*{\fill}\\
}\hlstd{\ \ \ \ }\hlstd{}\hlkwa{$<$biblioteca\textunderscore musical$>$}\hlstd{\hspace*{\fill}\\
}\hlstd{\ \ \ \ \ \ }\hlstd{}\hlkwa{$<$xsl:apply{-}templates select=}\hlstr{"/biblioteca\textunderscore musical/disco"}\hlkwa{/$>$}\hlstd{\hspace*{\fill}\\
}\hlstd{\ \ \ \ \ \ }\hlstd{}\hlkwa{$<$xsl:text$>$}\hlstd{\hspace*{\fill}\\
\hspace*{\fill}\\
}\hlstd{\ \ \ \ \ \ }\hlstd{}\hlkwa{$<$/xsl:text$>$}\hlstd{\hspace*{\fill}\\
}\hlstd{\ \ \ \ }\hlstd{}\hlkwa{$<$/biblioteca\textunderscore musical$>$}\hlstd{\hspace*{\fill}\\
}\hlstd{\ \ }\hlstd{}\hlkwa{$<$/xsl:template$>$}\hlstd{\hspace*{\fill}\\
}\hlstd{\ \ }\hlstd{}\hlkwa{$<$xsl:template match=}\hlstr{"/biblioteca\textunderscore musical/disco"}\hlkwa{$>$}\hlstd{\hspace*{\fill}\\
}\hlstd{\ \ \ \ }\hlstd{}\hlkwa{$<$xsl:apply{-}templates select=}\hlstr{"autor"}\hlkwa{/$>$}\hlstd{\hspace*{\fill}\\
}\hlstd{\ \ \ \ }\hlstd{}\hlkwa{$<$xsl:text$>$}\hlstd{\hspace*{\fill}\\
\hspace*{\fill}\\
}\hlstd{\ \ \ \ }\hlstd{}\hlkwa{$<$/xsl:text$>$}\hlstd{\hspace*{\fill}\\
}\hlstd{\ \ \ \ }\hlstd{}\hlkwa{$<$xsl:apply{-}templates select=}\hlstr{"titulo"}\hlkwa{/$>$}\hlstd{\hspace*{\fill}\\
}\hlstd{\ \ \ \ }\hlstd{}\hlkwa{$<$xsl:text$>$}\hlstd{\hspace*{\fill}\\
\hspace*{\fill}\\
}\hlstd{\ \ \ \ }\hlstd{}\hlkwa{$<$/xsl:text$>$}\hlstd{\hspace*{\fill}\\
}\hlstd{\ \ }\hlstd{}\hlkwa{$<$/xsl:template$>$}\hlstd{}\hspace*{\fill}\\
\hlkwa{$<$/xsl:stylesheet$>$}\hlstd{}\hspace*{\fill}\\
\mbox{}
\normalfont
\caption{One of the Type 2 XSL logicsheets found as solution by the
 algorithm. In fact, this solution was found 4 out of the 5 times it was run.\label{fig:sol}}
\end{figure}
\begin{table}[htbp]
\small
\begin{center}
\begin{tabular}{|l|c|c|c|}
\hline
 & Fitness & Time  \\
\hline
 Type 1  &  2.5 $\pm$ 0.8  &   240069 $\pm$ 42167  \\
 Type 2  &  2.4 $\pm$ 1.1   &   236556 $\pm$ 18430   \\
 Type 3  &  10.77823     &  269057 $\pm$ 79014 \\
\hline
\end{tabular}
\end{center}
\caption[]{Results for the second input document; in this case, success
 rate was 0 for all of them, so we show the average fitness of the best
 individual in the 100th generation, and running time.}
\label{tab:doc2}
\end{table}
The second input document is more complex, and, in fact, 100 generations
 are not enough to find a solution; however, once again Type 1 and 2 are
 more successful, achieving an average minimum fitness of around 2.5
 (optimum is close to 0, and is actually related to the minumun number
 of lines in the XSLT file divided by $10^5$), and doing it in around 4 minutes; Type 3 needs half
 a minute more (on average) to reach the same number of generations, but
 results are worse than the other two types of templates; running time
 graphs are shown in figure \ref{fig:time}.

\section{Conclusions and Future Work}
\label{sec:conc}

In this paper we present preliminary results of genetic programming
applied to XSLT logicsheets, as opposed to Lisp S-Expressions or other
type of programs; one of the advantages of this application is that
resulting logicsheets can be used directly in a production environment,
without the intervention of a human operator; besides, it tackles a
real-world problem found in many organizations.

In these initial experiments we have found which kind of XSLT template
structure is the most adequate for evolution, namely, one that matches
the \textsf{select} attribute in \textsf{apply-templates} with the
\textsf{match} attribute in templates, and an indeterminate number of
value-of instructions within each template. By constraining evolution
this way, we restrict the search space to a more reasonable size, and
avoid the high degree of degeneracy of the problem, with many different
structures yielding the same result, that, if combined, would result in
invalid structures. In general, we have also proved that a XSLT
logicsheet can be found just from an input/output pair of XML
documents. 
 
However, there are some questions  and issues that will have to be addressed in 
future papers:

\begin{itemize}
\item Using the DTD (associated to a XML file) as a source of information
for conversions between XML documents and for restrictions of the possible
variations.
\item Adding different labels in the XSLT to allow the building of 
different kinds of documents such as HTML or WML.
\item Considering the use of advanced XML document comparison tools (i.e. XMLdiff).
\item Analyzing different XSLT processors. Current application uses
Xalan but Saxon or XT might be faster.
\item Testing evolution with other kind of tools, such as a chain of SAX
      filters.
\item Obviously, testing different kinds and increasingly complex set of
      documents, and using several input and output documents at the
      same time, to test the generalization capability of the
      procedure. 
\end{itemize}


\bibliographystyle{plain}
\bibliography{xml,xml2,geneura,xslt-evolution} 

\end{document}